\documentclass[10pt,twocolumn,letterpaper]{article}

\usepackage{wacv}
\usepackage{times}
\usepackage{epsfig}
\usepackage{graphicx}
\usepackage{amsmath}
\usepackage{amssymb}
\usepackage[ruled]{algorithm2e}

\wacvfinalcopy 

\ifwacvfinal
\def\assignedStartPage{1} 
\fi

\ifwacvfinal
\usepackage[breaklinks=true,bookmarks=false]{hyperref}
\else
\usepackage[pagebackref=true,breaklinks=true,colorlinks,bookmarks=false]{hyperref}
\fi

\ifwacvfinal
\else
\fi

\begin{document}

\title{AdvFoolGen: Creating Persistent Troubles for Deep Classifiers}

\author{Yuzhen Ding\\
Arizona State University\\
{\tt\small yding58@asu.edu}
\and
Nupur Thakur\\
Arizona State University\\
{\tt\small nsthaku1@asu.edu}
\and
Baoxin Li\\
Arizona State University\\
{\tt\small baoxin.li@asu.edu}
}

\maketitle

\begin{abstract}
Researches have shown that deep neural networks are vulnerable to malicious attacks, where adversarial images are created to trick a network into misclassification even if the images may give rise to totally different labels by human eyes. To make deep networks more robust to such attacks, many defense mechanisms have been proposed in the literature, some of which are quite effective for guarding against typical attacks. In this paper, we present a new black-box attack termed AdvFoolGen, which can generate attacking images from the same feature space as that of the natural images, so as to keep baffling the network even though state-of-the-art defense mechanisms have been applied. We systematically evaluate our model by comparing with well-established attack algorithms. Through experiments, we demonstrate the effectiveness and robustness of our attack in the face of state-of-the-art defense techniques and unveil the potential reasons for its effectiveness through principled analysis. As such, AdvFoolGen contributes to understanding the vulnerability of deep networks from a new perspective and may, in turn, help in developing and evaluating new defense mechanisms.
\end{abstract}

\section{Introduction}

Deep neural networks have found wide applications in many computer vision tasks like face and object recognition, image segmentation, scene understanding etc., often delivering state-of-the-art performance for a given task. However, in recent years, it has been discovered that deep networks can be easily fooled/attacked: images can be created to trick a network into misclassification, although such created images may be classified correctly by humans. This has become a major concern for deep networks since they are becoming the backbone of real-world applications like access control and surveillance, where there may be adversarial agents constantly trying to outsmart the system.

Adversarial attacks on deep networks are broadly categorized as white-box, gray-box and black-box attacks, depending on the degree of access that the attacker has to the targeted network. The white-box attack is the one where the attacker has full access to the network (its architecture as well as the parameters). For a gray-box attack, the attacker has no access to the parameters but knows the architecture of the network. A black-box attack does not assume knowledge of a network's architecture or its parameters.

Although white-box attacks generally do better in baffling a targeted network than black-box attacks, the latter require minimum information of the targeted network and thus are favored by real-world attackers. Black-box attacks usually rely on a property called transferability \cite{liu2016delving} of the adversarial images for their design. This property refers to crafting the adversarial images via a substitute model and then using them for fooling the targeted model (since the parameters and the architecture of the targeted network is unknown). This can be explained considering the observation that the decision boundaries for a given feature space are similar for various networks if they all can deliver a high classification accuracy on the same test data.

Images generated using different attack types can be categorized as adversarial images or fooling images. Fooling images may look like random noise to human eyes but are given a class label with high confidence by a deep network. 
On the other hand, adversarial images may look just like some of the authentic images although often perturbations (either visible or imperceptible to human eyes) have been introduced to trick a network into misclassification. 

While many earlier attack algorithms can deliver high fooling ratio in fooling typical deep classifiers, recent years have seen effective defenses techniques \cite{das2018shield,raghunathan2018certified,dhillon2018stochastic,vijaykeerthy2018hardening,prakash2018deflecting,weng2018evaluating,xie2017mitigating}, which can defeat many existing attacking schemes. 
A defense mechanism can be as simple as retraining the network with adversarial images as additional inputs \cite{szegedy2013intriguing}. More complex mechanisms often employ a new network for explicitly identifying the adversarial images \cite{lu2017safetynet,song2017pixeldefend,samangouei2018defense}. Yet, new attack schemes keep emerging \cite{poursaeed2018generative,xiao2018generating,song2018constructing,yao2019trust,zheng2019distributionally}, and this attack-defense game will continue.

In this paper, we present a robust black-box attack approach that does not make explicit use of a substitute model for generating the images. This does not confine our attack to the group of the classifiers to which the substitute model belongs. Our approach employs a VAE-GAN \cite{larsen2016autoencoding} architecture as a building block and utilizes multiple target labels for constructing the objective function. These, in conjunction with employing a combination of both real and random images as input, help to produce new images lying in various portions of the original image feature space, hence creating a hurdle for existing defense mechanisms. As a result, our approach, termed as AdvFoolGen, can maintain a good fooling ratio in the face of many defense mechanisms. We call the created images as `AdvFool images' since they are not random noise but also not exactly similar to the original images either. 

We first evaluate our framework using initial fooling ratio as a criterion, compared with well-established attack approaches. We then showcase how these images are robust to existing defenses like retraining the network, adversarial training \cite{madry2017towards} and use of input transformations for retraining the network \cite{guo2017countering}. Moreover, we also illustrate why AdvFoolGen can fool the targeted network consistently through a principled analysis. As such, the work contributes to understanding the vulnerability of deep networks from a new perspective and may, in turn, help developing and evaluating new defense mechanisms.  

Section 2 discusses important related work in the area of adversarial attacks and defenses. Section 3 is an elaborate explanation of our approach. Section 4 contains the experiments and results. Section 5 includes the reasons why AdvFool images can fool the neural network even if state-of-the-art defenses are used, and Section 6 concludes the paper.

\section{Related Work}

The existence of the adversarial negatives of neural networks, despite the high performance of the network, was first discussed in \cite{szegedy2013intriguing}. \cite{nguyen2015deep} showed that it is extremely easy to fool a network with images that look like random noise to the human eye, crafted using evolutionary algorithms. There are two reasons mentioned for why these images get classified with high confidence. First, the fooling images contain the features matching one of the target class, because the evolutionary algorithms produce the features which are unique to a class rather than features from all the classes. Second, the neural networks do not learn the global structure of the objects but learn low-level and middle-level features. 

\cite{goodfellow2014explaining} presented a simple approach called Fast Gradient Sign Method (FGSM) for producing adversarial images which look similar to the original ones. It is a white-box attack that works by adding or subtracting the sign of the gradient such that the loss increases. They also mentioned that the linearity of the networks in high-dimensional spaces is the reason for the existence of such adversarial images. \cite{moosavi2016deepfool} proposed an attack called DeepFool based on iterative linearization of the target network for generating adversarial images. These images have minimum perturbations when compared to those produced by the FGSM method. 

The Carlini-Wagner (C$\&$W) attack, a white-box attack that produces very strong adversarial images, was first introduced in \cite{carlini2017towards}. It uses box-constraints and perturbation norm to form a cost function, which is optimized to generate adversarial images. Though the fooling ratio is 99\% for C$\&$W attack, they are computationally very expensive. In \cite{poursaeed2018generative}, an encoder-decoder architecture is used for generating the perturbations which are then added to the original image to form the adversarial image. This attack has a high fooling ratio and is faster than the previous adversarial attacks. To generate adversarial images robust to the input transformations, \cite{athalye2017synthesizing} proposed an algorithm called Expectation over Transformation. 

More recently, there are works utilizing variants of Generative Adversarial Networks (GANs) for generating adversarial images to design fast and efficient attacks \cite{xiao2018generating,song2018constructing,zhao2018generating}. These methods assume that they either have access to the target classifier (semi-white box) or take a substitute distilled model (black-box) when training the generator of GAN which generates the adversarial examples. However, such assumption may not be realistic in real scenarios. Moreover, the adversarial examples generated by above approaches have been narrowed down to a relatively small subspace of the space where the original images lie.
 
To tackle the hazard caused by the adversarial/fooling images, several methods for defending against these adversarial attacks have been proposed. Initially, \cite{szegedy2013intriguing} mentioned that training the network on a combined dataset of original and adversarial images improves the robustness of the network to adversarial attacks. \cite{madry2017towards} further discussed the adversarial robustness of the deep neural networks using MNIST and CIFAR-10 datasets. It also presents an iterative strategy called adversarial training. Here, the network is trained on adversarial images generated during the training stage. Another effective defense, known as defensive distillation, was proposed in \cite{papernot2016distillation}. Based on the use of the knowledge from a DNN for training another neural network, this defense produced classifiers which are less sensitive to perturbations.  

\cite{guo2017countering} proposed the use of input transformations like JPEG compression, image quilting, bit-depth reduction, total variance minimization on the images before retraining to achieve higher robustness. A faster way of defending against adversarial images was proposed in \cite{xie2017mitigating}. The input images are passed through a random resizing and padding layer before feeding it to the neural network. This eliminates the need for retraining the network, which saves time and resources. 

\section{Proposed Approach: AdvFoolGen}

Consider a classification network $P$ trained on clean images of $C$ different classes, and the space of the normalized natural images $X_o$ is represented by $\mathbb{R}^{[0,1]}$. Without loss of generality, we assume $P$ achieves good performance on the clean images. Our aim is to generate adversarial or/and fooling images $X_{af}$ that belong to the same space as the clean images but can achieve high fooling ratio even with various defense mechanisms applied. More specifically, if the correct class of an original image $x_o$ is $c$, our goal is to make the pre-trained network $P$ predict the corresponding adversarial or/and fooling image $x_{af}$ anything other than $c$, with the constraint of the distance between $x_o$ and $x_{af}$ to be as small as possible. Note that we only test on network $P$ and require no access to the parameters or gradients from $P$. 
This framework can also be extended to other applications like segmentation with appropriate changes in the similarity distance and loss. Without losing generality, we focus on the task of fooling the image classification models.

\begin{figure*}[h]
    \centering
    \includegraphics[width=0.9\linewidth,height = 5.0cm]{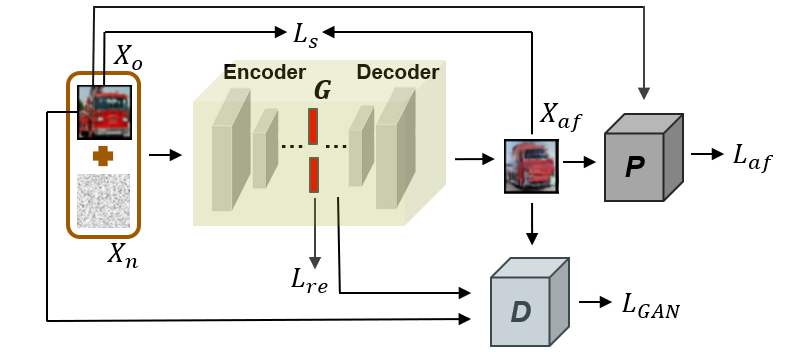}
    \caption{The proposed AdvFoolGen framework for generating AdvFool images. The input is a 4-channel image consisting of three original image channels and a noise channel. The discriminator network $D$ and pre-trained network $P$ compute their respective losses with the original image and the corresponding AdvFool image as inputs.}
    \label{fig:model_structure}
\end{figure*}

Although the feature space $\mathbb{R}^{[0,1]}$ of $X$ is almost infinite in terms of the various combinations of different values, the original image set itself only occupies a relatively small region in $\mathbb{R}^{[0,1]}$ \cite{liu2016delving}. This leaves a huge space for the attackers to explore for success. On the other hand, the adversarial or/and fooling images generated by the current attack algorithms often gather in another small region in $\mathbb{R}^{[0,1]}$, which in turn leaves room for a defense scheme to work by enabling the target network $P$ to recognize the small, `poison' region where majority of the attacking images are present. We propose a new framework for generating diverse and robust adversarial/fooling images by forcing the generated images to utilize the feature space $\mathbb{R}^{[0,1]}$ greedily. Moreover, the min-max game training strategy is used for training a Generative Adversarial Network (GAN) to strengthen the generator in our attack model.

\begin{algorithm}
\SetAlgoLined
\SetKwData{Left}{left}\SetKwData{This}{this}\SetKwData{Up}{up}
\SetKwFunction{Union}{Union}\SetKwFunction{FindCompress}{FindCompress}
\SetKwInOut{Input}{Input}\SetKwInOut{Output}{Output}
\Input{Original images $X_o$, Noise mask $X_n$, model $G$, $D$ and $P$}
\Output{AdvFool image $X_{af}$}
\For{each epoch $e = 1,2,...$}{
\While{Training}{
\For{each batch $b = 1,2,...$}{
Generate a noise mask $X_n$\;
Construct input image as $ cat(X_o,X_n)$\;
\For{$i = 1$ \KwTo $5$}{\label{forins}
Update model D\;}
Update model G\;}}
\While{Testing}{
Generate a noise mask $X_n$\;
Construct input image as $ cat(X_o,X_n)$\;
Output $X_{af}$\;
Compute fooling ratio using the predicted label given by $P$\;}}
\caption{AdvFoolGen}
\label{model_algorithm}
\end{algorithm}

Fig. \ref{fig:model_structure} illustrates the architecture of our framework. The input is a four-channel image $x_i$ consisting of an original colored image $x_o$ and a gray-scale noise image of one channel $x_n$, with a relative magnitude. $x_i$ is first fed to an encoder to learn the parameters of the latent distribution $\mu_{x}$ and $log(\sigma_{x})$, followed by a re-sampling process from the learned $\mu_{x}$ and $log(\sigma_{x})$. Next, the re-sampled latent representation goes through a decoder to reconstruct an image $x_{af}$. Note that in our case, the structure of the encoder and the decoder is not symmetric as the input consists of four channels and the reconstructed image consists of three channels only. The next step is to feed both $x_o$ and $x_{af}$ to a discriminator $D$ to detect whether the image is real or fake. Simultaneously, $x_o$ and $x_{af}$  are passed through the pre-trained model $P$ to check if the predicted labels are different. We followed the WGAN \cite{arjovsky2017wasserstein} training strategy to ensure stability of training. The loss is defined by four components. The first one is the re-sampling loss $L_{re}$ (Eq. (\ref{loss_re})). The parameters of the latent representation are drawn from a multivariate Gaussian distribution $N(0,1)$. 

\begin{equation}
    L_{re} =D_{KL}(N(\mu_x,\sigma_{x})||N(0,1))
    \label{loss_re}
\end{equation}
where $D_{KL}$ represents the KL divergence. 

The second loss component is the similarity loss $L_{S}$ (Eq. (\ref{loss_S})), which aims at decreasing the pixel-wise distance between $x_{o}$ and $x_{af}$.

\begin{equation}
    L_{S} = ||x_o-x_{af}||_{2}
    \label{loss_S}
\end{equation}

The third loss component is the GAN loss $L_{GAN}$ (Eq. (\ref{loss_GAN})) that uses a discriminator to distinguish $x_{af}$ from $x_{o}$ and in turn makes the generator stronger.

\begin{equation}
    L_{GAN} = log(D(x_o))+log(1-D(G(z)))
    \label{loss_GAN}
\end{equation}

The last loss component is the fooling loss $L_{af}$ (Eq. (\ref{loss_advfool})). Based on the type of an attack (non-targeted attack or targeted attack), the fooling loss takes different forms. In our work, we force the predicted label of image $x_{af}$ to be close to the two least likely classes simultaneously, unlike the previous fooling losses which force the predicted label to be one target label only \cite{goodfellow2014explaining,carlini2017towards,moosavi2016deepfool}. 

\begin{equation}
    L_{af} = -\sum_{i = 1}^{2}log(H(P(x_{af})),1_{t_i})
    \label{loss_advfool}
\end{equation}
where $H(\cdot)$ is the cross-entropy function, and $t_i$ represents the $i$-th target label. It is worth mentioning that this fooling loss can force the adversarial images to lie close to the decision boundary of the two least likely classes in the feature space, which causes more confusion for the classifier even with defenses applied. We discuss more details about it in Section 5.   

The total loss for the AdvFoolGen is the weighted summation of the above four losses, given by Eq. \ref{loss_total}. The entire algorithm is summarized in Algorithm \ref{model_algorithm}.
\begin{equation}
\begin{split}
    &L = \alpha L_{re}+\beta L_{S}+\gamma L_{GAN}+\lambda L_{af}\\
    &subject~ to \quad  \alpha+ \beta+\gamma+\lambda=1. 
    \end{split}
    \label{loss_total}
\end{equation}

\begin{figure} [h]
    \centering
    \includegraphics[width=0.96\linewidth,height = 3.5cm]{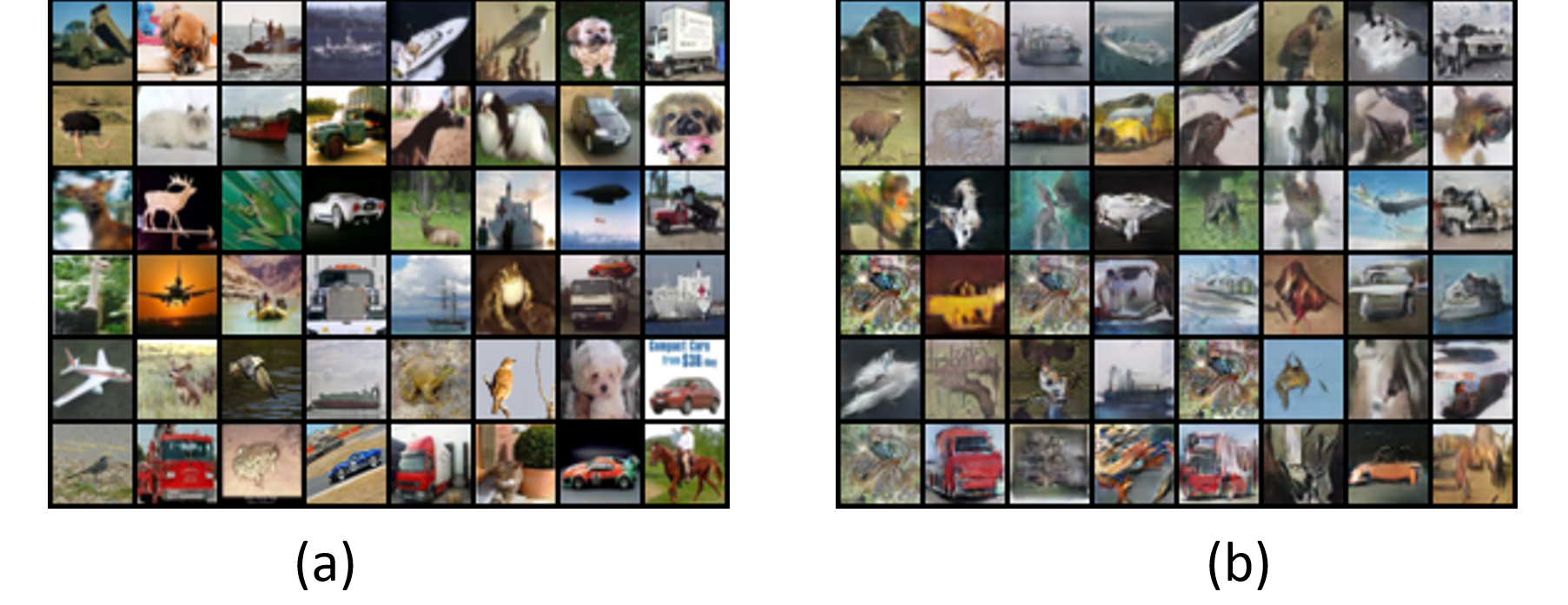} 
    \caption{(a) The original images from CIFAR-10 dataset. (b) The corresponding AdvFool images. Though the difference between the AdvFool and original images is visible to human eye, these images still capture the colors and object patterns from the original images.}
    \label{fig:samples_ori_adv}
\end{figure}

In the existing adversarial image generators, one critical constraint on the adversarial image is resemblance to the original image. In our case, we impose a relaxed version of such constraint because of the following considerations: 1) This constraint is a subjective one because the similarity threshold for the images may vary from person to person, making it an inconsistent criterion; 2) It confines the adversarial images to a small subspace which can be easily defended. To utilize the feature space greedily, our generator generates images that lie between the adversarial images and fooling images. Thus, we name the images as AdvFool images. The AdvFool images do not look like noise since it contains some patterns from the corresponding original image. On the other hand, the difference is obvious to the human eye. Fig. \ref{fig:samples_ori_adv} shows some AdvFool images along with their corresponding original images from the CIFAR-10 dataset. In our experiments, we found that the AdvFool images yielded a better fooling ratio even with various defenses applied. The details are presented in the next section. 

\section{Experiments}

This section includes the experimental results showing the success of our approach. First, the experiments are compared with well-established baseline attacks - Fast Gradient Sign Method (FGSM) \cite{goodfellow2014explaining}, Iterative Fast Gradient Sign Method (I-FGSM) \cite{kurakin2016adversarial}, DeepFool \cite{moosavi2016deepfool}, Carlini $\&$ Wagner (C$\&$W) \cite{carlini2017towards} and Generative Adversarial Perturbations (GAP) \cite{poursaeed2018generative} in terms of the initial fooling ratio explicitly. Next, we show the robustness of the AdvFool images by re-attacking the target network strengthened with state-of-the-art defense strategies like retraining the network, adversarial training \cite{madry2017towards}, bit-depth reduction and JPEG compression \cite{guo2017countering}. We also include the details about the experimental settings and evaluation criteria in this section. 

\subsection{Experimental Setting}

\begin{table*} [h]
  \centering
    \label{initialfool}
  \begin{tabular}{| c|c|c|c |} 
  \hline
    \bf Attack Algorithm   &\multicolumn{3}{|c|}{\bf Initial Fooling Ratio} \\ \cline{2-4}
    & \bf CIFAR10 & \multicolumn{2}{|c|}{\bf TinyImagenet}\\
    \cline{3-4}
   & & \bf Top1 & \bf Top5\\
    \hline\hline
    FGSM   &92.82\% * &88.55\% * &75.18\% * \\
    \hline
    I-FGSM     &99\% *  &100\% * &98.86\% *\\
    \hline
   DeepFool     &99\% &99\% &83.77\% \\
    \hline
    C$\&$W    &100\%&99.12\%  &90.63\%  \\
    \hline
    GAP    &82\% &94.98\%   &87.01\% \\
    \hline
  AdvFoolGen    & 68.5\% - 78.36\%**  &95.41\%-97.65\%**  & 90.14\%-93.07\%**\\
    \hline
  \end{tabular}
  \caption{Initial fooling ratio for the AdvFoolGen compared with state-of-the-art attacks on CIFAR-10 and TinyImageNet Dataset. *$\epsilon = 0.07$ for FGSM and I-FGSM. **We report a range for AdvFoolGen attack as the fooling ratio varies from epoch to epoch.}
\end{table*}

We use CIFAR-10 dataset \cite{krizhevsky2014cifar} and TinyImageNet dataset \cite{le2015tiny} for our experiments. CIFAR-10 contains 10 classes, and each class has 5,000 training images and 1,000 test images. Each image has a height and width of 32 and consists of 3 channels. The target network is VGG-19 \cite{simonyan2014very} classifier trained on clean CIFAR-10 images achieving a test accuracy of 92.42\%. TinyImagenet is a smaller version of ImageNet dataset. It has 200 classes, and each class has 500 training and 50 validation images. Each images size is $64 \times 64 \times 3$. The target network for TinyImagenet is ResNet18 with Top1 and Top5 validation accuracy of 72.3\% and 91.2\% respectively. No overfitting is observed while training the classifiers for both datasets. We use the test and validation data to report our results for CIFAR-10 and TinyImageNet respectively. 

For the AdvFool images generator, the noise mask $x_n$ is drawn from a uniform distribution of $U(0,mgn)$, where $mgn$ is the noise magnitude. Though the choice of noise magnitude is not unique, we found that different magnitudes yield similar performance through an empirical study. Thus, we used a fixed value of 0.1. The model is trained from scratch and it reaches a relatively stable stage after a few epochs. Thereafter, we use the AdvFool images generated from different epochs to attack the pre-trained network.

To demonstrate the robustness of AdvFool images, we attack the networks equipped with defenses. It is worth mentioning that initially, we set the ground truth label for each $x_{af}$ to be $t_c$, the true label of the corresponding original image $x_o$. However, the AdvFool images are visually between original images and fooling images, and thus using the original labels is not fair for the defense. Therefore, we assign additional labels for $x_{af}$ when applying defenses that require ground truth of AdvFool images. More information about the labeling of AdvFool images is discussed in the later part of this section.

\subsection{Initial Fooling Ratio}

To evaluate the success of the attack, we use the fooling ratio as the evaluation metric which is calculated using Eq. (\ref{fooleq}).
\begin{equation}
    F_{ratio} = \frac{\sum_{c=0}^{C}t_{c}^{x_{af}}\neq t_{c}}{N}
    \label{fooleq}
\end{equation}
where $t_{c}^{x_{af}}$ represents the predicted label for $x_{af}$, $t_c$ is the predicted label for $x_{o}$ and $N$ is the number of all images in the test set.

Table \ref{initialfool} shows the initial fooling ratio of different attack methods on CIFAR-10 dataset and TinyImagenet. AdvFoolGen achieves a reasonable (more than half the images can fool) but not a competitive (15\%-30\% lower than others) fooling ratio as compared to the existing approaches. As TinyImageNet dataset is a smaller version of ImageNet dataset, to remain consistent with the notations the latter dataset uses, we report two fooling ratios - Top1 and Top5. It can be seen that Top1 fooling ratio is always higher than the Top5 fooling ratio, which is intuitive since the Top1 accuracy is always lower than the Top5 accuracy. The moderate Top1 accuracy of the target classifier explains the high Top1 and Top5 fooling ratio for all the attacks. Similar to the results obtained on CIFAR-10 dataset, we observe that the fooling ratio of AdvFoolGen attack is lower than most of the state-of-the-art attacks. However, the majority of the AdvFool images can successfully fool the network. 

\begin{table*}[h]
  \centering
    \label{reattack}
  \begin{tabular}{|c|c|c|c|c|c| } 
  \hline
    \bf Attack     & \bf Retraining*  & \bf Adv Training  &\bf BDR-3   & \bf BDR-8  &\bf JPEG   \\
    \hline
    \hline
    FGSM   &9.76\% &35.9\%  & 18.21\%  &16.2\%    &18.6\%\\
    \hline
    I-FGSM     &8.22\% &39.3\%  & 12.32\%  &11.2\%   &13.1\%\\
    \hline
   DeepFool     &9.87\% & 26.5\%  & 14.55\%  &14.1\%  &14.8\%   \\
    \hline
    C$\&$W    &9.2\% &41.25\%   & 12.97\%  &12.19\%   &15.67\%  \\
    \hline
    GAP    &8.91\% &9.04\%  & 14.99\%  &15.09\%   &19.89\%  \\
    \hline
 AdvFoolGen    & \bf 27.3\%-58.1\%  & \bf 59.56\%-65.26\%   & \bf 37.08\%-52.82\%   &\bf24.76\%-35.4\%   & \bf 24.44\%-50.64\%   \\
    \hline
  \end{tabular}
   \caption{Fooling ratio after the defenses are applied. The fooling ratio for AdvFoolGen is higher than existing attacks when it comes to networks with added defense mechanisms. For Bit-Depth Reduction (BDR), the results are reported for bit-depth of 3 and 8. For JPEG, all the images are compressed at the quality level of 75 (out of 100). *Equal number of original and adversarial images are used for retraining. For AdvFoolGen attack, 5000 AdvFool images are used for training and 1000 AdvFool images are used for testing and the total number of classes is 11.}
\end{table*}

Although the initial fooling ratio is a common measure to evaluate how good an attacker is, it has several drawbacks. Using just the fooling ratio, we cannot evaluate the diversity of the generated images nor robustness of the attack, which are two critical properties contributing to consistent fooling in real-world applications. In other words, if an attacker can only fool the pre-trained network once and fails under one simple defense technique, it is not a strong attacker. Thus, the adversarial/fooling images that can fool the pre-trained network strengthened with defenses consistently have drawn more attention recently. In the next section, we demonstrate the robustness of AdvFool images with experiments using various defenses.

\subsection{Effect of Defenses on Fooling Ratio}

Table \ref{reattack} shows the fooling ratio of different attacks as the defenses are employed for CIFAR-10 dataset. The results on TinyImageNet are included in the supplementary material. The $2^{nd}$ column shows the fooling ratio against retrained networks. For the existing attackers, we retrain the pre-trained network with the same number of the adversarial images as that of the original images. For AdvFoolGen, we retrain the target network by slightly revising the structure: more output labels instead of 10, the details of which are discussed next. 

As AdvFool images are between adversarial and fooling images, we tested them using three different retraining strategies. First, we used their original class labels i.e. treating them as adversarial images for retraining. Second, we created 10 new corresponding classes for the AdvFool images i.e. if the original label of an AdvFool image is 2, then it will be assigned label 12 while retraining. In this case, we get a total of 20 classes. Lastly, we created a new class for all the AdvFool images making the total number of classes equal to 11. For the first two types of retraining, the network could not learn the AdvFool images very well (around 25\% test accuracy on AdvFool images) and eventually led to more confusion, reducing the accuracy on original images significantly. However, for the 11-class retraining strategy, the network could learn the AdvFool images very well (around 95\% test accuracy on AdvFool images). Therefore, we present all the results for AdvFoolGen attack with one additional class (11-class strategy). If the network does not classify an AdvFool image as belonging to $11^{th}$ class, it is considered to be fooling the network.

Compared with baseline attacks, although all the fooling ratio decreased significantly on the retrained networks, at least 30\% of the AdvFool images still fooled the retrained network. Also, the accuracy on the original images only slightly decreased to $88.05\%$. 

Considering the $3^{rd}$ column, it is observed that the adversarially trained networks can no longer be fooled by the state-of-the-art attacks with a high fooling ratio. However, AdvFool images can fool such a network effortlessly. Moreover, we observed that with increased number of training AdvFool images from different epochs, the accuracy on both original and AdvFool images decreased significantly. The high fooling ratio indicates that AdvFoolGen can get past the adversarial training strategy with ease. 

\begin{figure}[h]
    \centering
    \includegraphics[width=0.8\linewidth]{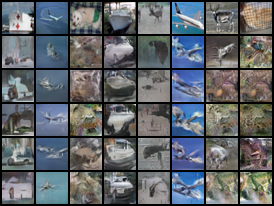}
    \caption{The first row shows the original images from CIFAR-10 dataset. Row 2-5 are AdvFool images generated using generators of different epoch. The variations between the images from different epochs are clearly visible. Though the images are morphed, they do not resemble objects from any other classes. }
    \label{fig:samplesfromdiffepoch}
\end{figure}

The $4^{th}$ and $5^{th}$ columns show the results for Bit-Depth Reduction on the input images followed by retraining. We use a bit depth of 3 and 8. The last column shows the JPEG compression defense results where all the images are compressed at the quality level of 75 (out of 100). Yet, a significant number of AdvFool images can fool the network retrained with the transformed images. 

\begin{figure*}[h]
    \centering
    \includegraphics[width=0.9\linewidth]{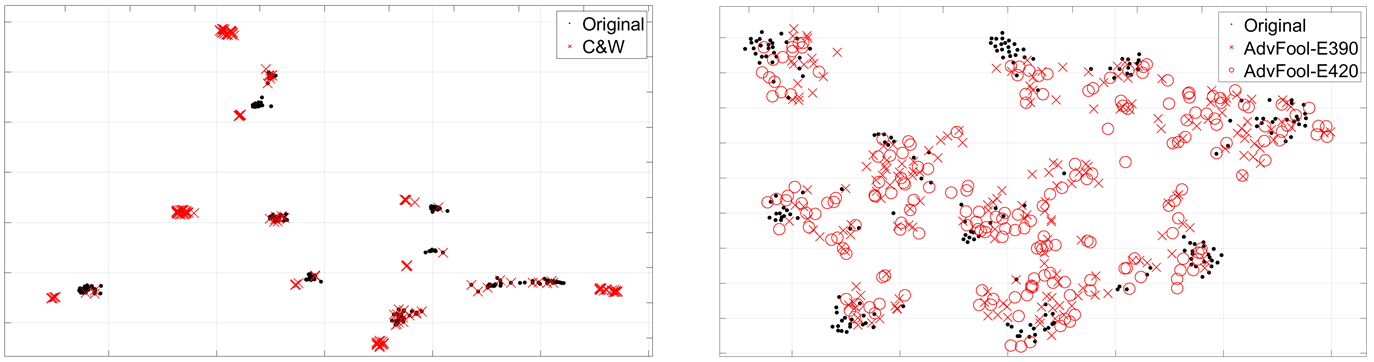}
    \caption{Visualizing the features from the layer before classification layer of VGG-19 network in 2-D for C$\&$W (left) adversarial CIFAR-10 images and the AdvFool (right) CIFAR-10 images. The corresponding original images are represented in both of the figures by black dots. Two sets of AdvFool images are shown, each from a different epoch generator. While the C$\&$W images remain clustered in a small subspace, the AdvFool images are spread out and between the classes explaining its higher fooling ratio for defense-equipped networks. Besides, the two sets of AdvFool images do not overlap completely explaining its fooling ability across various epochs.}
    \label{cw_advfool}
\end{figure*}

The fooling ratio for all the baseline attacks drop dramatically after the defenses are applied. A decrease in the fooling ratio is observed for the AdvFoolGen attack as well but it is relatively high compared to the baseline attacks. AdvFoolGen attack suffers only 20\%-30\% decrease in the fooling ratio. On an average, 30\% of the AdvFool images still fool the network. Therefore, in real-world applications, AdvFool images can be considered as more `poisonous' than the adversarial images generated by existing attacks. 

Taking a closer look at Table \ref{reattack}, we found that the I-FGSM and C$\&$W attacks with almost 100\% initial fooling ratio, completely failed when simple defenses were employed. On the contrary, FGSM which has the lowest fooling ratio among the baseline approaches achieves high fooling ratio against the defenses. This supports our claim that the initial fooling ratio may not be a good criterion for evaluating the adversarial attacks. Furthermore, the difference between the initial fooling ratio and re-attack fooling ratio implies that the adversarial images generated by C$\&$W attack overfit the target network. Thus, these images fail immediately with only a few changes to the target network.

\section{Why AdvFool Images can Fool the Network?}

In this section, we unveil the reasons why AdvFool images can fool the network in the face of various defense techniques using two different perspectives. We first carefully examine the architecture of our model to discover potential factors leading to the effectiveness and robustness of AdvFool images. Next, we use statistical tools to analyze AdvFool images from various epochs to verify several conjectures that explain the good fooling ratio of AdvFool images. 

\subsection{Revisiting the architecture of AdvFoolGen}

\begin{figure}
    \centering
    \includegraphics[width=1\linewidth,height = 3.5cm]{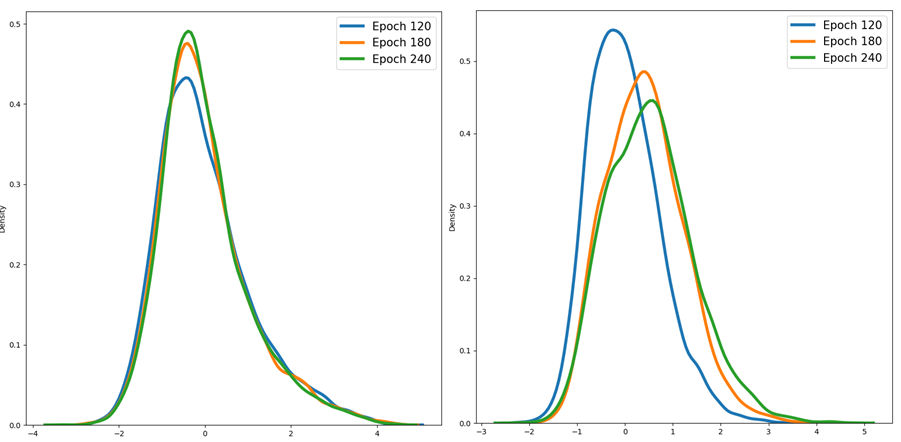}  
    \caption{The distributions of the mean (left) and variance (right) used for latent representation of AdvFool images in three different epochs of the AdvFoolGen attack. For each epoch, the distribution is Gaussian with different parameters.}
    \label{fig:mean_logvar_distribution}
\end{figure}

From Fig. \ref{fig:model_structure}, it is clear that the reference image of an AdvFool image is not a pure original image, but a 4-channel image which is an integration of the original colored image and an added channel of gray-scale noise image of relatively small magnitude. We introduce the noise to bring in randomness in the input which helps the generator to explore the feature space deeper. Although the noise magnitude has to be chosen from a limited range, the variations of the value within this limited range are unlimited. The re-sampling step adds extra randomness which forces the generator to explore the untouched regions as well. Fig. \ref{fig:samplesfromdiffepoch} shows some AdvFool images from different epochs. It is clear that the images from different epochs have different perturbations which are visible to human eye. Due to the almost unlimited range of noise magnitude and the re-sampling process, every single time a different set of AdvFool images are generated. Therefore, the network can be fooled even if images from different epochs are used for retraining. Besides, as the number of AdvFool images used for retraining increase, the accuracy on the original images starts to decrease which is highly undesirable.

\begin{table*}[h]
\centering
\label{kld}
\begin{tabular}{ |c|c|c|c|c| } 
\hline
\bf Epoch &\multicolumn{2}{c|}{\bf Mean} &\multicolumn{2}{c|}{\bf Variance}\\
\cline{2-5}
&{$\bf KLD (P\| \bf Q)$}  & $\bf KLD (Q\| \bf P)$   & $\bf KLD (P\| \bf Q)$   &$\bf KLD (Q\| \bf P)$\\
\hline
\hline
{120} \&{180} &{$4.91\times 10^{10}$} &{$1.236\times10^{10}$} &{$1.35\times10^{12}$} &{$2.99\times10^{11}$}\\
\hline
{120} \&{240} &{$10.26\times10^{10}$} &{$1.298\times10^{10}$} &{$2.513\times10^{12}$} &{$3.135\times10^{11}$}\\
\hline
{120} \&{330} &{$10.309\times10^{10}$} &{$1.299\times10^{10}$} &{$2.539\times10^{12}$} &{$3.136\times10^{11}$}\\
\hline
{120} \&{360} &{$10.313\times10^{10}$} &{$1.30\times10^{10}$} &{$2.546\times10^{12}$} &{$3.1368\times10^{11}$}\\
\hline
\end{tabular}
\caption{KL-Divergence between the distributions of the mean and variance of the latent representation. $P$ and $Q$ are the first and second epoch number mentioned in a particular row in the first column.}
\end{table*}

We defined fooling loss $L_{af}$ in such a way that the AdvFool images should be on/near the boundary region of two least-likely classes, where the original images rarely occur. The relaxed constraint of the similarity between the original image and AdvFool image along with the discriminator network ensures that the AdvFool image does not visually go too far away from the corresponding original image. As a result of these constraints in the fooling loss, the AdvFool images are morphed but never completely change into images that belong to any of the remaining classes. Thus, these images lie between adversarial images and fooling images, making it hard for the classification model to determine their correct labels required for retraining the network. Moreover, the AdvFool images capture features from three different classes (the original class and two least-likely classes). It tends to confuse humans while labeling resulting in inconsistent labels. Therefore, human labeling for AdvFool images is not feasible.  

\subsection{Analytical Study of AdvFool Images}

To further support our claims about the AdvFool images, we analyze the AdvFool images systematically using statistical tools. Fig. \ref{cw_advfool} is the visualization of the deep features extracted for a typical batch of CIFAR-10 dataset by the VGG-19 network just before the classification layer in 2-D. The original images (black dots) are well-clustered in both the figures. In the first figure, the C$\&$W adversarial images are present in a small subspace which is in the opposite direction of the original image categories. Therefore, these images can be easily defended with defenses like retraining. However, it is clear from the second figure that the AdvFool images are quite spread out, lying near the decision boundary of two classes. Also, the set of AdvFool images from two different epochs do not overlap which makes defending them difficult. 

In the re-sampling process of AdvFoolGen, the values of mean and variance are sampled assuming that the distribution is Gaussian. We claim that the AdvFool images from different epochs can fool the network because they belong to different distributions. To verify this claim, we use density estimation and show that the mean and variance sampled for the latent distribution at different epochs belong to different distributions. We used a non-parametric density estimation technique called Parzen-Window with Gaussian window function to estimate the distribution of the mean and variance. The smoothing parameter for this technique is found using Grid Search. 

As mean and variance from the re-sampling process have high dimensions, we used Principal Component Analysis (PCA) to reduce their dimensions for visualization and illustration purpose. The dimensionality reduction is done in such a way that all of them are mapped to the same dimension to make a fair comparison. Fig. \ref{fig:mean_logvar_distribution} shows the distributions of the mean (left) and variance (right) used in the re-sampling process during 3 different epochs. It is evident that all of these are Gaussian distributions but with different parameters. Though the difference between the distributions look very minute in the figures, it is enough to result in distinct distributions that can produce different set of AdvFool images. Such sets of AdvFool images can fool the network successfully. It becomes difficult to defend these images which are generated from different epoch generators as they belong to different distributions. In other words, even if the network learns AdvFool images from one epoch, it can be easily fooled by another set of AdvFool images from a different epoch as they do not belong to the same distribution. 

We calculated the KL-Divergence between two different distributions to show that even a tiny difference in the parameters leads to very distinct distributions. Table \ref{kld} lists the KL-Divergence between the distributions from different epochs. It is clear that the KL-Divergence increases as we move farther away from a particular epoch number. This justifies the increase in the fooling ratio as a farther epoch generator is used for generating the AdvFool images. For example, if a network is familiar with the AdvFool images from epoch 120, the AdvFool images from epoch 360 will achieve high fooling ratio than those from epoch 180. As KL-Divergence is not symmetric, we calculated it in the reverse manner as well and observed the same trend. Therefore, the order in which the KL-Divergence is calculated does not make an impact on the results. 

\section{Conclusion}

In this paper, we proposed a new black-box scheme, AdvFoolGen, for attacking deep classifiers. AdvFoolGen generates images (termed as AdvFool images) which can consistently fool a network that has the help of many existing defense techniques. From experiments, we observed that the initial fooling ratio is not a good metric for evaluating an attack scheme, as the scheme may fall apart under some defense mechanisms. Though AdvFoolGen has relatively lower initial fooling ratio, it can keep deceiving the defense strategies over and over again. 

The principled analysis shows that the mean and variance used for the latent representation of the images in our framework belong to different Gaussian distributions for different epochs. Thus, the images from AdvFoolGen at a certain epoch can fool the network trained on images produced from the AdvFoolGen at another epoch. The analysis also showed that the AdvFool images do not occupy a small subspace and are highly spread. Also, sets of AdvFool images from different epoch generators do not overlap in space explaining why they can fool the network after defenses are applied using a set of AdvFool images. The success of our attack on the defenses shows the susceptibility of the current defense mechanisms and raises a need for more robust DNNs. It provides a better understanding of the pitfalls of the neural networks which is useful for building more generic and advanced defense mechanisms in the future. 

\begin{table*} [h]
\scriptsize
  \centering
  \begin{tabular}{ c|c|c|c|c} 
  \hline
    \bf Attack Algorithm     & \bf Retraining*  & \bf Adversarial Training &\bf BDR-3  &\bf JPEG    \\
    \hline\hline
    FGSM   &30.8\% &49.37\%  & 51.92\%     &54.18\%  \\
    \hline
    I-FGSM     &40.5\% &48.74\%  & 48.44\%    &51.15\% \\
    \hline
  DeepFool     &29.2\% &47.36\%  & 43.02\%  &47.76\%  \\
    \hline
    CW    &30.04\% &48.61\%   & 46.95\%   &47.26\%   \\
    \hline
    GAP    &34.09\% &33.76\%  & 33.55\%   &35.21\%   \\
    \hline
  AdvFoolGen**    &\bf  43.1\%-57.2\%  &\bf 54.6\%-61.0\%  & \bf 40.3\%-66.4\%   &\bf  42.1\%-63.9\%   \\
    \hline
  \end{tabular}
  \caption{Top 1 fooling ratio after the defenses are applied on TinyImageNet dataset. The fooling ratio for AdvFoolGen is higher than existing attacks when it comes to networks with added defense mechanisms. For Bit-Depth Reduction, a bit-depth of 3 is used. *Equal number of original and adversarial images are used for retraining. For AdvFoolGen attack, 500 AdvFool images are used for training and 50 AdvFool images are used in validation set as a new class is added for them. **We report a range for AdvFoolGen attack as the fooling ratio varies from epoch to epoch.}
    \label{reattack_TINY_top1}
\end{table*}

\begin{table*}  [h]
\scriptsize
  \centering
  \begin{tabular}{ c|c|c|c|c } 
  \hline
    \bf Attack Algorithm     & \bf Retraining*  & \bf Adversarial Training &\bf BDR-3   &\bf JPEG    \\
    \hline\hline
    FGSM   &18.26\% &22.28\%  & 27.12\%   &26.78\%  \\
    \hline
    I-FGSM     &17.28\% &20.96\%  & 20.26\%   &23.09\% \\
    \hline
  DeepFool     &16.07\% &15.26\%  & 20.96\%    &18.98\%  \\
    \hline
    CW    &14.35\% &16.71\%   & 18.55\%    &18.88\%   \\
    \hline
    GAP    &12.81\% &11.91\%  & 12.77\%    &13.3\%   \\
    \hline
  AdvFoolGen**    & \bf 24.8\%-33.2\%  &\bf 28.9\%-35.2\%  & \bf 20.1\%-32.4\%   & \bf 20.6\%-35.6\%   \\
    \hline
  \end{tabular}
  \caption{Top 5 fooling ratio after the defenses are applied on TinyImageNet dataset. The fooling ratio for AdvFoolGen is higher than existing attacks when it comes to networks with added defense mechanisms. For Bit-Depth Reduction, a bit-depth of 3 is used. *Equal number of original and adversarial images are used for retraining. For AdvFoolGen attack, 500 AdvFool images are used for training and 50 AdvFool images are used in validation set as a new class is added for them. **We report a range for AdvFoolGen attack as the fooling ratio varies from epoch to epoch.}
    \label{reattack_TINY_top5}
\end{table*}

{\small
\bibliographystyle{ieee_fullname}
\bibliography{egbib}
}

\section{Supplementary Material}

In this supplementary material, we provide additional results to support our claims proposed in the main submission, AdvFoolGen. The additional results provided are on TinyImageNet dataset. We show reattack Top1 and Top5 fooling ratio on networks equipped with different defenses.

\section{Additional Experimental Results}
In this section, we provide the additional results obtained on TinyImageNet dataset for AdvFoolGen attack. 

\subsection{Effect of Defenses on Fooling Ratio}

Table \ref{reattack_TINY_top1} and Table \ref{reattack_TINY_top5} show the Top1 and Top5 reattack fooling ratio on TinyImageNet dataset for different attacks, respectively. The target network here is strengthened with effective defenses. The fooling ratio varies for different epochs for AdvFoolGen and therefore a range of fooling ratio is reported. In this case too, Top1 fooling ratio is higher than the Top5 fooling ratio. Though there is a decrease in the fooling ratio to some extent after the defenses are used for all the attacks including AdvFoolGen, it is clear that the decrease in fooling ratio of AdvFoolGen is comparatively lower. 

The fooling ratio achieved by different attack algorithms on retrained target networks is shown in Column 2. For the existing attack algorithms, equal number of adversarial and original images are used for retraining. For the reasons mentioned in the main submission, we use a network with one additional class for the AdvFool images while retraining. As each class contains 500 training images and 50 validation images in TinyImageNet dataset, we use 500 AdvFool images for training and 50 AdvFool images for validation. It is seen that all the state-of-the-art attacks fail to fool the retrained network with high fooling ratio but almost half of the AdvFool images can still fool it. 

The next column presents the fooling ratio when the target network is adversarially trained. As the number of AdvFool images from generators at different epoch increase in the training set, the accuracy on original as well as AdvFool images decrease. All other attacks we compare with can be easily defended using adversarial training.

The last two columns are the defenses which use transformed images for retraining in order to defend against adversarial attacks. The transformations like Bit-Depth Reduction and JPEG compression are applied to the images before using them for retraining the network. Column 4 displays the results for Bit-Depth Reduction transformation with a Bit-Depth of 3. The last column is the defense which uses JPEG compressed adversarial images for retraining the network. The average fooling ratio of AdvFoolGen attack for both these defenses is comparable to FGSM, but outperforms all other attacks. This demonstrates that the AdvFool images can fool the network regardless of the type of image transformation applied.

Carefully examining the results obtained, it is observed that the attacks with high initial fooling ratio experience a significant decrease in the fooling ratio after the defenses are applied. This low fooling ratio shows that the existing attacks are not strong adversarial attacks and can be defended with small changes in the network. The AdvFoolGen attack is stronger than the existing ones because it can fool the networks equipped with state-of-the-art defenses.

\end{document}